\documentclass[wcp]{jmlr}

\usepackage{longtable}% for long tables

\usepackage{booktabs}
\usepackage{xcolor}
\usepackage{listings}
\usepackage{textcomp}

\definecolor{codegreen}{rgb}{0,0.6,0}
\definecolor{codegray}{rgb}{0.5,0.5,0.5}
\definecolor{codepurple}{rgb}{0.58,0,0.82}
\definecolor{backcolour}{rgb}{0.95,0.95,0.92}

\lstdefinestyle{mystyle}{
    backgroundcolor=\color{backcolour},   
    commentstyle=\color{codegreen},
    keywordstyle=\color{magenta},
    numberstyle=\tiny\color{codegray},
    stringstyle=\color{codepurple},
    basicstyle=\ttfamily\footnotesize,
    breakatwhitespace=false,         
    breaklines=true,                 
    captionpos=b,                    
    keepspaces=true,                 
    numbersep=5pt,                  
    showspaces=false,                
    showstringspaces=false,
    showtabs=false,                  
    tabsize=2
}
\def\BibTeX{{\rm B\kern-.05em{\sc i\kern-.025em b}\kern-.08em
    T\kern-.1667em\lower.7ex\hbox{E}\kern-.125emX}}
\par

\makeatletter
\let\Ginclude@graphics\@org@Ginclude@graphics 
\makeatother

\title{Deep Learning Models on CPUs: \\A Methodology for Efficient Training}

  \author{\Name{Quchen Fu$^1$} \Email{quchen.fu@vanderbilt.edu}\\
  \addr Dept. of Computer Science, Vanderbilt University, Nashville, TN, USA
  \AND
  \Name{Ramesh Chukka} \Email{ramesh.n.chukka@intel.com}\\
  \Name{Keith Achorn} \Email{keith.achorn@intel.com}\\
  \Name{Thomas Atta-fosu} \Email{thomas.atta-fosu@intel.com}\\
  \Name{Deepak R. Canchi} \Email{deepak.r.canchi@intel.com}\\
  \addr Intel Corporation, Santa Clara, CA
  \AND
  \Name{Zhongwei Teng} \Email{zhongwei.teng@vanderbilt.edu}\\
  \Name{Jules White} \Email{jules.white@vanderbilt.edu}\\
  \Name{Douglas C. Schmidt} \Email{d.schmidt@vanderbilt.edu}\\
  \addr Dept. of Computer Science, Vanderbilt University, Nashville, TN, USA
 }

\begin{document}

\maketitle

\begin{abstract}
GPUs have been favored for training deep learning models due to their highly parallelized architecture. As a result, most studies on training optimization focus on GPUs. There is often a trade-off, however, between cost and efficiency when deciding how to choose the proper hardware for training. In particular, CPU servers can be beneficial if training on CPUs was more efficient, as they incur fewer hardware update costs and better utilize existing infrastructure. This paper makes several contributions to research on training deep learning models using CPUs.  First, it presents a method for optimizing the training of deep learning models on Intel CPUs and a toolkit called ProfileDNN, which we developed to improve performance profiling. Second, we describe a generic training optimization method that guides our workflow and explores several case studies where we identified performance issues and then optimized the Intel\textsuperscript{\textregistered} Extension for PyTorch, resulting in an overall 2x training performance increase for the RetinaNet-ResNext50 model. Third, we show how to leverage the visualization capabilities of ProfileDNN, which enabled us to pinpoint bottlenecks and create a custom focal loss kernel that was two times faster than the official reference PyTorch implementation.
\end{abstract}

\section{Introduction}
Deep learning (DL) models have been widely used in computer vision, natural language processing, and speech-related tasks (\cite{mattson2020mlperf}~\cite{shen2023fishrecgan}~\cite{jiang2020deep}~\cite{wu2023rethinking}). Popular DL frameworks include PyTorch (\cite{paszke2019PyTorch}), TensorFlow (\cite{abadi2016tensorflow}), and OpenVINO (\cite{gorbachev2019openvino}), etc. The hardware can range from general-purpose processors, such as CPUs and GPUs, to customizable processors, such as FPGA and ASICs, that are often called XPUs (\cite{reinders2021sycl}).
\footnotetext[1]{Work performed during an internship at Intel, data in this paper are intentionally reported as relative to comply with Intel Policy}

All these varieties of hardware make it hard to propose a universal methodology for the efficient training of DL models. Since GPUs have dominated  deep learning tasks, comparatively little attention has been paid to optimizing models running on CPUs, especially for training (\cite{kalamkar2020optimizing}). Previous DL model research conducted about CPUs focused mostly on performance comparison of CPUs and GPUs (\cite{wang2019benchmarking}; \cite{buber2018performance}; \cite{shi2016benchmarking}; \cite{dai2019benchmarking}), or only focused on CPU inference (\cite{qian2020profiling}). 

A key question to address when optimizing training performance on CPUs is what metrics should guide the optimization process (\cite{shen2023learning}). Several metrics and benchmarks have been proposed to measure DL workload and training performance. For example, Multiply-Accumulate (MAC) has been used as a proxy for flops to measure computational complexity for Convolutional Neural Network (CNN) models (\cite{chang2018reducing}). Time-to-Train (TTT) has been widely adopted to measure the training performance of a DL model by measuring the time models take to reach certain accuracy metrics.
NetScore (\cite{wong2019netscore}) was proposed as a universal metric for DL models that balances information density and accuracy. Until recently, however, no widely accepted benchmark for DL models existed that incorporated a wide range of domain tasks, frameworks, and hardware.

 MLPerf (\cite{mattson2019mlperf}) was proposed as a comprehensive DL benchmarking suite that covers a variety of tasks and hardware. Many major tech companies have contributed to this effort by competing for better performance. Intel has been actively participating in the MLPerf challenge to improve the training performance of Deep learning models across multiple domains. 
 
 To address portability issues related to AI running on different hardware platforms, Intel has open-sourced the oneAPI Deep Neural Network Library (oneDNN) (\cite{OneDNN}), which is a cross-platform performance library of basic deep learning primitive operations, including a benchmarking tool called \textit{benchDNN}. Intel has also created optimized versions of popular frameworks with oneDNN, including Intel\textsuperscript{\textregistered} Optimizations for TensorFlow and Intel\textsuperscript{\textregistered} Extensions for PyTorch (\cite{Ipex}). Few guidelines exist, however, for profiling and optimizing DL model training on CPUs. 
 
 Several fundamental research challenges must be addressed when training DL models on CPUs, including the following:
\begin{enumerate}
    \item \textbf{How to locate bottlenecks}. Since frameworks with CPU-optimized kernels (such as Intel\textsuperscript{\textregistered} Extention for PyTorch) are relatively new, generic model-level (\cite{torres2021computational}) profilers (such as the PyTorch Profiler(\cite{paszke2019PyTorch})) are not oneDNN-aware. Moreover, low-level profilers like \textit{benchDNN} can only benchmark performance at the operational level. Specifically, identifying primitive operations most critical for specific model/framework/hardware combinations is essential so that low-level (e.g., oneDNN level) optimizations can optimize performance significantly.
    \item \textbf{How to fix bottlenecks}. While GPUs have well-established platforms (such as CUDA (\cite{cuda})) for kernel implementations, Deep Neural Network Libraries for CPU are less well-known. It is therefore essential to understand how to fix performance bottlenecks, e.g., by locating and implementing custom operation kernels for both forward and backward propagation, as well as adopting proper low-precision training so computing time can be reduced without sacrificing accuracy for CPUs.
    \item \textbf{How to set achievable goals}. Projections for CPUs are often done in a crude way by dividing CPU performance in flops over flops required for model training. In a computation-bounded scenario, it is essential to create an experiment-based projection for deep learning models so that the goal is realistically achievable, i.e., not only theoretically achievable but also considers hardware limits and kernel optimizations.

\end{enumerate}
To address these challenges, we designed a structured top-down method that helped us prioritize different optimizing options for training DL models (e.g., RetinaNet (\cite{lin2017focal})) on CPUs. Incorporating this new approach, we also developed a DL performance profiling toolkit called ProfileDNN that is oneDNN-aware and supports profiling and projection at the model level, thereby bridging the gap for oneDNN-specific model-level projection.

The remainder of this paper is organized as follows: Section~\ref{subsec:profile} and Section~\ref{subsec:datadis} summarize different profile tools and their contribution to locating hot spots and discrepancies; Section~\ref{subsec:proj} describes projection goal and procedure, as well as ProfileDNN's structure and workflow; Section~\ref{subsec:dl} through Section~\ref{subsec:custom_k} discuss recommendations and approaches to enable efficient training without sacrificing accuracy; Section~\ref{subsec:distributed} analyzes the training efficiency and convergence under distributed situation; and Section~\ref{subsec:future} presents concluding remarks and our future work. All  experiments in this paper were performed on Intel Xeon Cooper Lake processors.

\section{Method Summary}

First, we summarize the method component of our contribution to optimizing training on CPUs. Our goal is to provide a structured approach for users to optimize training DL models on CPUs. Our method adopts a top-down approach similar to what \cite{yasin2014top} has described that aims to \textit{locate the critical bottlenecks in a fast and feasible manner}.

\begin{figure}[hpbt]
\centering
\vspace{-0.1in}
\includegraphics[width=0.7\linewidth]{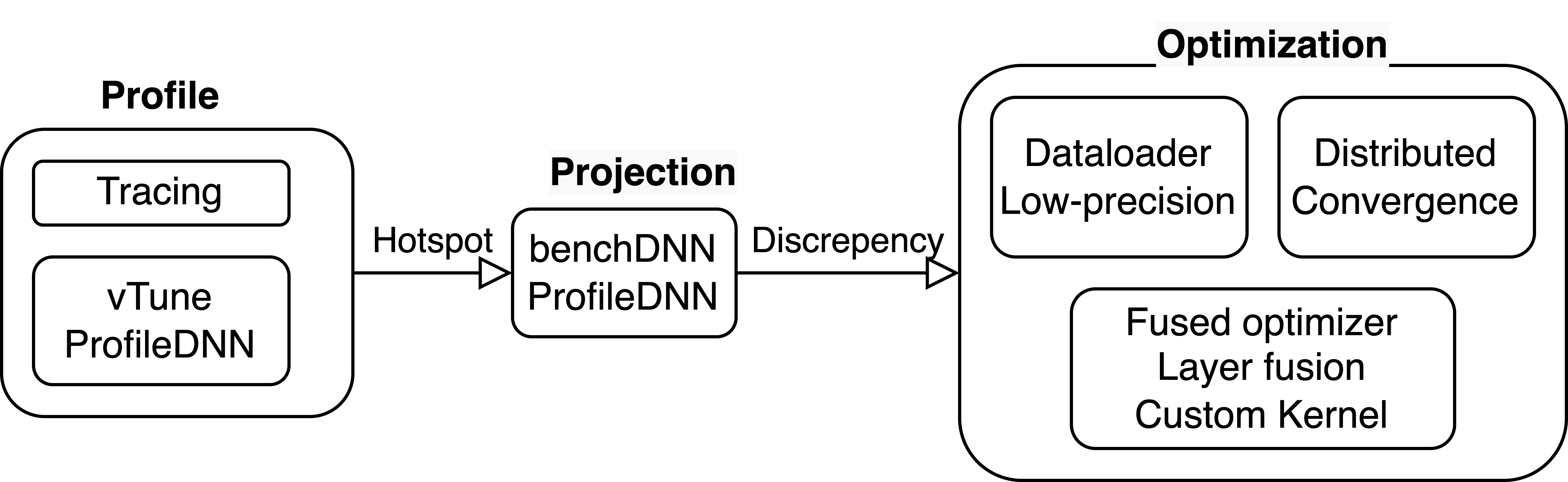}
\caption{Method Flowchart}
\label{fig:method}
\vspace{-0.15in}
\end{figure}

We believe the workflow can be roughly categorized into three stages: profiling, projection, and optimization. As shown in Fig \ref{fig:method}, we break each stage into different tile groups.
Users are advised to follow the method groups from left to right, as each group benefits from the previous group's results. Our toolkit ProfileDNN can work both as a profile tool and a projection tool. 

\subsection{Profile and Tracing}
\label{subsec:profile}
During the profile stage, users should observe the breakdown of operation kernel components of the DL model and their relative significance. Special attention should be paid to the discrepancies between their model and data versus the reference implementation and original use case. For example, do all the major kernel operations of reference exist in their model? Likewise, does the kernel component percentage stays about the same? If the answer to either question is "no" their code may perform worse due to poor oneDNN kernel adoption.

ProfileDNN helps users better compare the distribution of the kernel components by producing intuitive visualization. This tactic was also adopted by vTune (\cite{reinders2005vtune}). ProfileDNN supports all primitive kernels \textit{(conv, pool, matmul, reorder, etc)} from \textit{benchDNN}.

\begin{figure}[hpbt]
\centering
\includegraphics[width=0.8\linewidth]{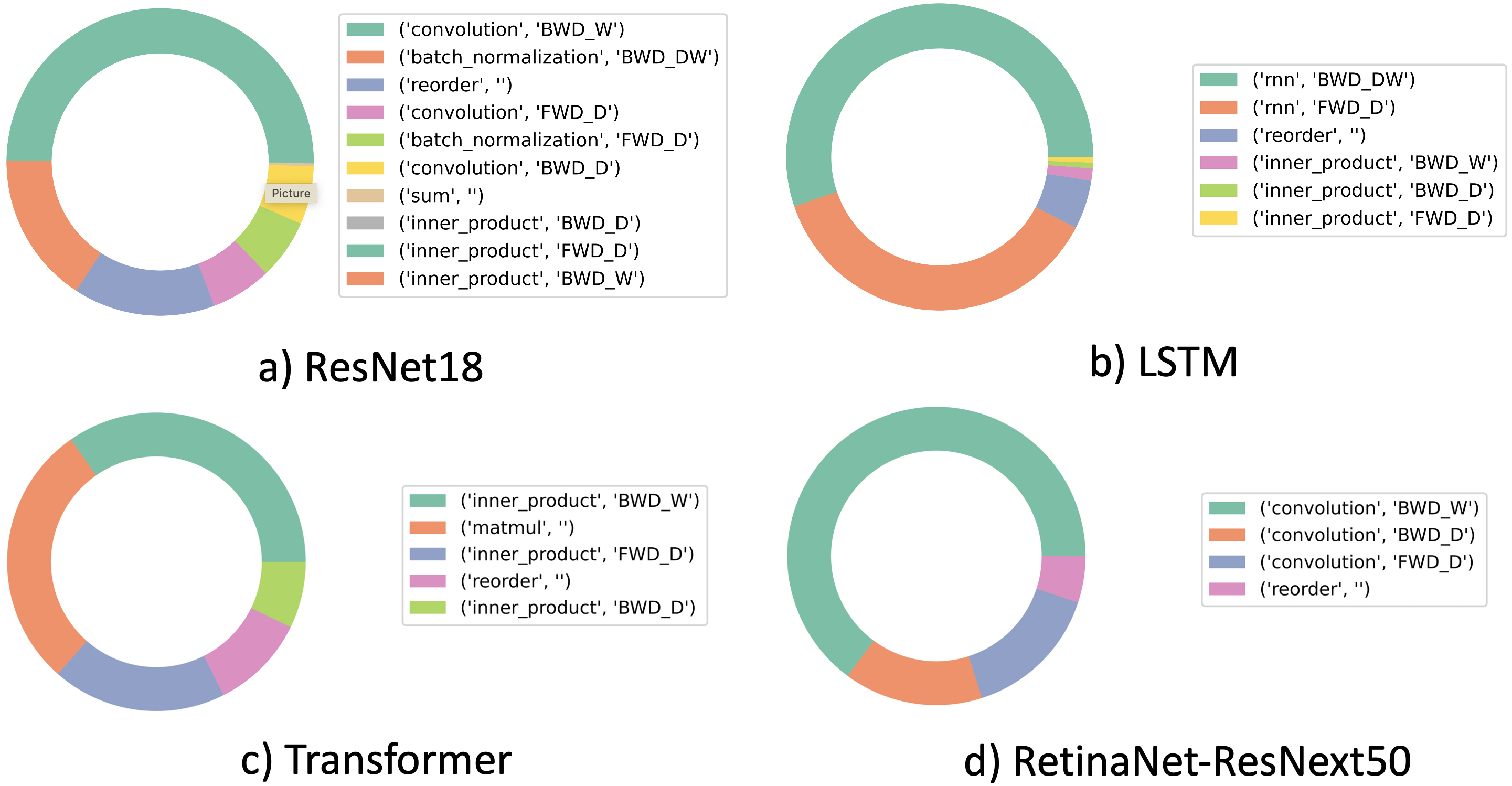}
\caption{Comparison of Primitive Operations Across Models}
\label{fig:comparemodel}
\end{figure}

Convolutional Neural Networks (CNNs) (\cite{krizhevsky2012imagenet}), Recurrent Neural Networks (RNNs) (\cite{graves2013speech}), and Transformers (\cite{vaswani2017attention}) are some of the most popular Neural Network models today. ProfileDNN can break down the primitive operations by type and directory, as shown in Fig~\ref{fig:comparemodel}a-c. We found that both CNN and RNN models spend more time doing back-propagation than forward-propagation. Transformer models consist mostly of inner product and matrix multiplication, which correspond to the \texttt{softmax} operation that is often a performance bottleneck for transformer-based models (\cite{lu2021soft}~\cite{li2023text}). 

Fig~\ref{fig:comparemodel}d also plots the breakdown of the RetinaNet-ResNext50 model, which is a complicated object detection model. The distribution in this figure looks similar to the CNN in Fig~\ref{fig:comparemodel}a.

A primitives-level breakdown is often sufficient to locate model bottlenecks since many DL training tasks are computation-bound. In the case of a memory/cache-bounded scenario, however, a tracing analysis is needed to inspect the orders in which each operation runs. A trace is an ordered set of span sequences, where each span has an operation name, a start and end timestamp, as well as relations to other spans (child process, etc). If a trace is highly fragmented there is significant context switching, so a custom merged operator may help improve performance.

\begin{figure}[hpbt]
\centering
\includegraphics[width=0.5\linewidth]{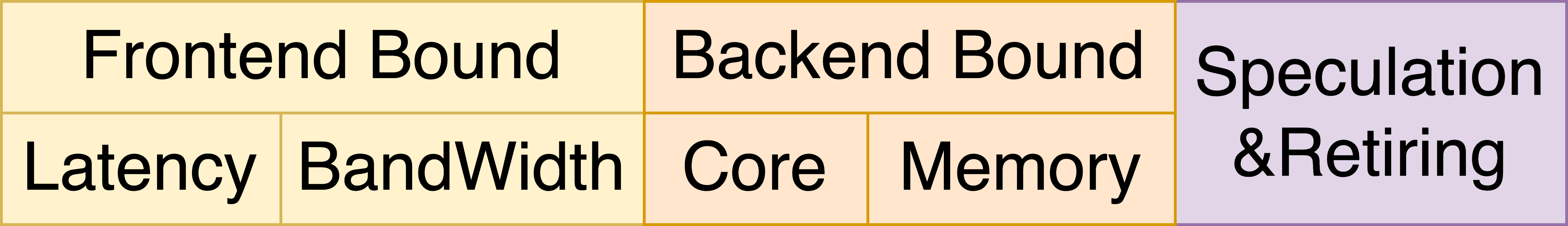}
\caption{The vTune Design}
\vspace{-0.14in}
\label{fig:vtune}
\end{figure}

VTune is another very powerful tool for profiling CPU and heavily adopts the top-down methodology (\cite{yasin2014top}). vTune divides the CPU workflow pipeline into frontend and backend, with the former bounded by latency and bandwidth, and the latter bounded by core (computation) and memory (cache), as shown in Fig~\ref{fig:vtune}. 
The first round of profiling should be a generic hot spot analysis on training the model to determine costly operations. 

The profiling round can be followed by micro-architecture exploration that measures CPU utilization rate (spinning time), memory bandwidth, and cache (L1, L2, or L3) miss rate. After pinpointing the primitive operation with the most computation-heavy footprint, algorithm- or implementation-level optimizations can be applied. If memory is the bottleneck, memory access and IO analysis can also be performed on individual operations. 

\subsection{Data Discrepancy}
\label{subsec:datadis}
An easily overlooked discrepancy is the difference between the reference dataset and the custom dataset. The data distribution can not only affect the performance of the same model, but it can also sometimes change the model itself (\cite{shen2023data}). For example, RetinaNet-ResNext50 is a classification model that changes structure based on the number of classes from the dataset. 

\begin{figure}[hpbt]
\centering
\vspace{-0.15in}
\includegraphics[width=0.65\linewidth]{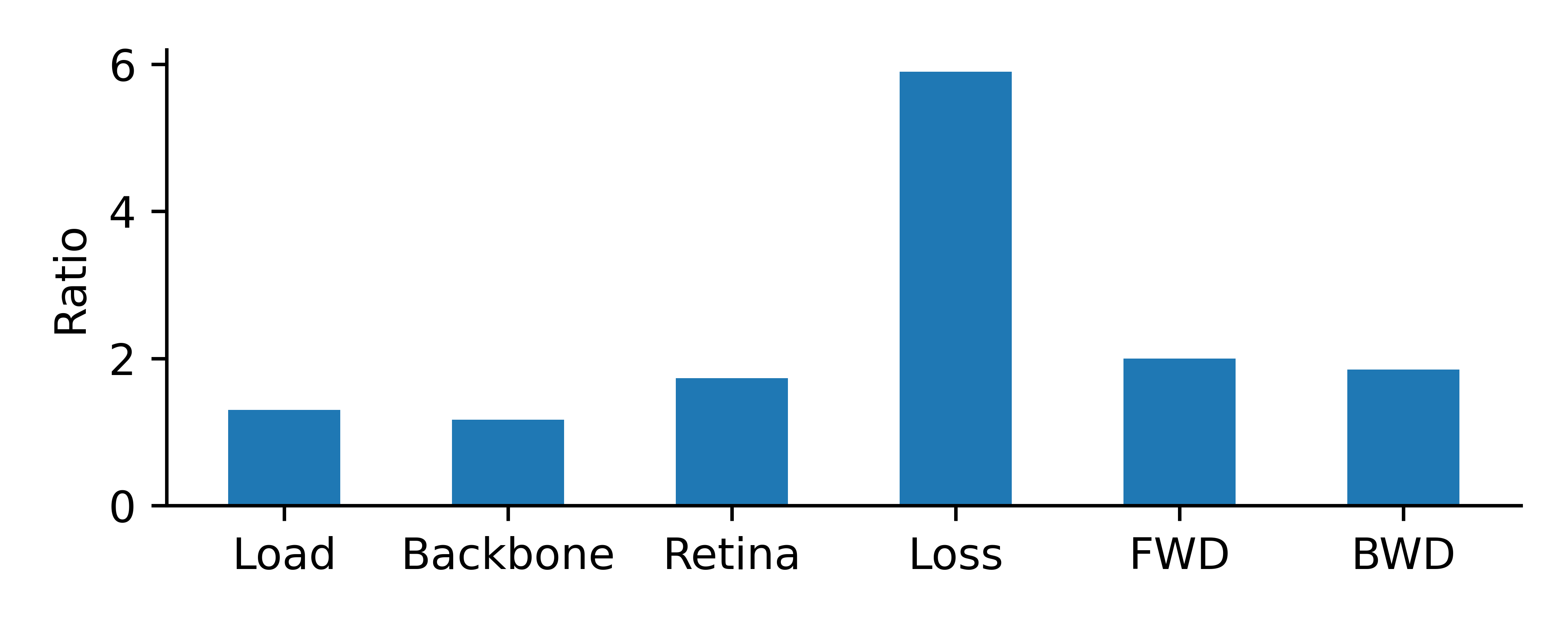}
\vspace{-0.1in}
\caption{Open Image vs COCO Training Time Ratio Breakdown}
\vspace{-0.15in}
\label{fig:ratio}
\end{figure}
After we switched the dataset from COCO (\cite{lin2014microsoft}) to OpenImage (\cite{kuznetsova2020open}), the training time increased dramatically. We found that the dataset size increased 10 times, but the training time per epoch increased 20 times, which is not proportional. Part of this increase can be attributed to a bigger fully connected (FC) layer in the backbone. In particular, we found that the major increase in time is within the focal loss calculation caused by three times more classes, as shown in the detailed breakdown in Fig \ref{fig:ratio}.

A tracing analysis also corresponded to the conclusion by showing that one-third of the backward calculation time was spent on focal loss. We addressed this issue by implementing our custom focal loss kernel, as discussed in Section \ref{subsec:custom}.

\subsection{Projection and Toolkit structure}
\label{subsec:proj}

The projection of DL models aims to determine the theoretical performance ceiling of a specific model/framework/hardware combination. Intel has an internal tool that can perform projection for DL models, but this tool requires much manual setting and tuning. \textit{BenchDNN} can be used to project performance on specific hardware automatically, but only one operation at a time. We therefore designed ProfileDNN to combine the advantage of both since it can project the whole DL model with little manual effort.

\begin{figure}[hpbt]
\centering
\vspace{-0.12in}
\includegraphics[width=0.55\linewidth]{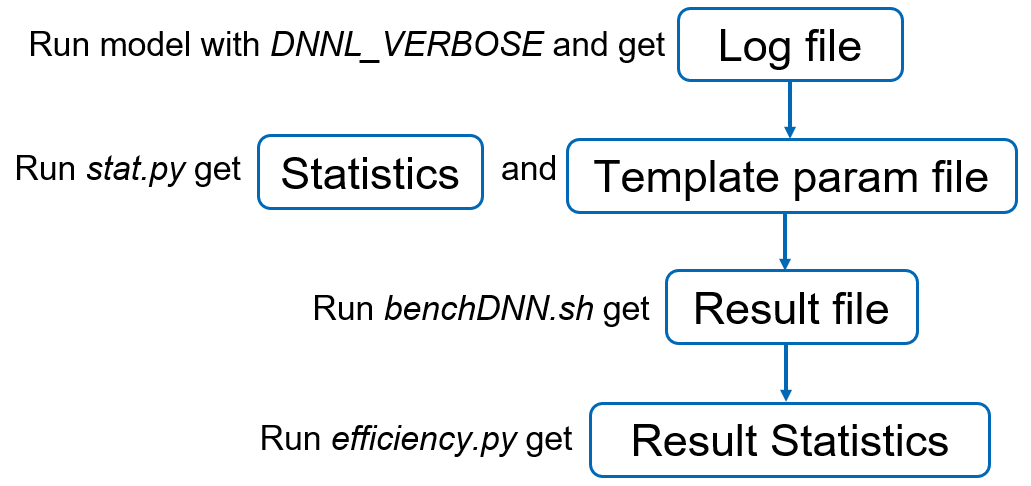}
\caption{Toolkit Structure and Flow Pipeline}
\vspace{-0.12in}
\label{fig:struct}
\end{figure}

As is shown in Fig~\ref{fig:struct}, ProfileDNN takes in an arbitrary log file produced by running deep learning models on a platform that supports oneDNN with \texttt{DNNL\_VERBOSE} set to \texttt{1}. The \textit{stats.py} file then collects and cleans the raw log file into CSV format, produces a template parameter file, calculates and plots the component distribution of primitive operations. The \textit{benchDNN.sh} file runs each primitive operation multiple times and takes the average. The \textit{efficiency.py} then takes a weighted sum of all operations' time by the number of calls and produces an efficiency ratio number. 

To ensure our toolkit can accurately reproduce the running behavior of the kernels from the original model, we ensure both the computation resources and the problem descriptions are the same. We use \textit{numactl} to control the number of CPU cores and memory binding and the \textit{mode} is set to \textit{p (performance)} in \textit{benchDNN} to optimize performance. These parameters are carefully controlled and summarized in Table \ref{table:benchDNN}.
\begin{table}[hbpt]
\begin{center}
\begin{tabular}{|c|c|c|}
\hline
\textbf{Name}  & \textbf{Example} & \textbf{Description}\\
\hline
Driver &	conv, relu, matmul, rnn, bnorm & -\\
\hline
Configuration &	u8s8u8, s8f32 &	Data type\\
\hline
Directory & FWD\_I, BWD\_D, BWD\_W &	Op specific\\
\hline
Post\_Ops & sum+eltwise\_relu & Optional\\
\hline
Algorithm & DIRECT & Op specific\\
\hline
Problem\_batchsize & mb1, mb32 & \\
\hline
Problem\_input & id4ih32iw32 & Op specific\\
\hline
Problem\_output & id16ih16iw16 & Op specific\\
\hline
Problem\_stride & sd2sh4sw4 & Op specific\\
\hline
Problem\_kernel & kd2kh3kw3 & Op specific\\
\hline
Problem\_padding & pd1ph1pw1 & Op specific\\
\hline
Problem\_channel & ic16oc32 & Op specific\\
\hline
\end{tabular}
\caption{\label{table:benchDNN}Summary of benchDNN Parameters}
\end{center}
\vspace{-0.5in}
\end{table}

\subsection{Dataloader and Memory Layout}
\label{subsec:dl}

By examining the DL training process from the same vTune top-down perspective shown in Fig \ref{fig:vtune}, the data loader can be seen as a frontend bounded by bandwidth and latency. There are three sources of bottlenecks for the data loader: I/O, decoding, and preprocessing. We found similar performance for data in NVMe or loaded to RAM and the I/O overhead is negligible. We observed a better decoding performance by adopting Pillow-SIMD and accimage as the backend in torchvision.

A PyTorch dataloader parameter controls the number of worker processes, which are usually set to prevent blocking the main process when training on GPUs. For training on CPUs, however, this number should not be set to minimize memory overhead. Since CPU RAM memory is usually larger than GPU memory---but has a smaller bandwidth---training on CPUs has the advantage of allowing larger batch size and training larger model (\cite{wang2019benchmarking}).

Here we define \textit{n} as batchsize, \textit{c} as channel, \textit{h} as height, and \textit{w} as width. The recommended memory layout in Intel\textsuperscript{\textregistered} Extension for PyTorch is \textit{nhwc} (channel last) for more efficient training, though the default layout in \textit{benchDNN} is \textit{nchw}. We set the default behavior of ProfileDNN to adopt \textit{nchw} to follow tradition. If the log input specifies the memory layout, ProfileDNN will automatically override the default.

\subsection{Library Optimization}
\label{subsec:opt}
Substituting slow operation implementations with a more efficient library can improve performance significantly, as we discovered by replacing the official PyTorch implementation with the Intel\textsuperscript{\textregistered} Extension for PyTorch counterpart. ProfileDNN helped identify a discrepancy between the number of backward convolution calls between the official PyTorch vs. the Intel\textsuperscript{\textregistered} Extension for PyTorch library. Using a detailed analysis of the computation graph and our ProfileDNN-based visualization, we found calls emanated from the frozen layers in the pre-trained model (ResNext backbone). 

Our analysis helped increase the performance of RetinaNet-ResNext50 model training  with 2 fixed layers by 16\%. We also found that the primitive operation \texttt{frozenbachnorm2d} was missing in Fig \ref{fig:comparemodel}d and \texttt{torchvision.ops.misc.FrozenBatchNorm2d} was interpreted as \texttt{mul} and \texttt{add} ops, which meant it was not a single oneDNN kernel operation. 

Our analysis indicated that bandwidth-limited operations made the \texttt{torchvision.ops.misc.FrozenBatchNorm2d} operation inefficient. It therefore cannot be fused with other operations to reduce  memory accesses. Training performance  increased by 29.8\% after we replaced the \texttt{torchvision.ops.misc.FrozenBatchNorm2d} operation with \texttt{IPEX.nn.FrozenBatchNorm2d}.

\subsection{Low-precision Training}
\label{subsec:quant}

Low-precision training has proven an efficient way for high-performance computing and BF16 (Brain Floating Point) is widely supported by various Deep learning hardware. BF16 is unique since it has the same range as float32 but uses fewer bits to represent the fraction (7 bits). This BF16 datatype characteristic can be beneficial when computing speed is important, but can also lead to accuracy loss when compared with float32 in calculating the loss. As shown in Fig \ref{fig:kernel}, computation time is almost half when done in BF16 compared to float32.

There is a significant discrepancy between the forward/backward training time ratio compared with that of bare-bone kernel time. This discrepancy indicates highly inefficient non-kernel code in the forward pass. We found that the loss function does not scale well and comprises a significant portion of computation time.

After locating the focal loss as having significant overhead, we implemented our version of the focal loss kernel. However, the loss result is different from the original implementation. We pinpointed the accuracy loss as happening during low-precision casting to BF16 by \texttt{torch.cpu.amp.autocast}. Unless convergence can be guaranteed, therefore, casting data into BF16 should be avoided for loss calculation, especially when reduction operations are involved.

\subsection{Layer Fusion and Optimizer Fusion} 

In inference mode, certain layers can be fused for a forward pass to save cache copying operation since an intermediate is not needed. In training mode, however, the layers containing trainable weights need to save the intermediate for backpropagation. When oneDNN is in inference mode, it enables \texttt{batchnorm+relu} and \texttt{conv+relu} respectively, but not \texttt{frozenbatchnorm (FBN)+relu}. 

OneDNN already supports \texttt{eltwise (linear, relu)} post-ops for \texttt{conv} and chaining of post-ops. We therefore treat \texttt{FBN} as a per-channel linear operation to enable \texttt{conv+FBN+relu}. This fusion potentially increases performance 30\% and is a work-in-progress (WIP). Intel\textsuperscript{\textregistered} Extension for PyTorch currently supports fusion of SGD (\cite{robbins1951stochastic}) and the Lamb (\cite{you2019large}) optimizer. We tested a fused/unfused Lamb optimizer with RetinaNet and found a 5.5X reduction in parameter updating time when the optimizer is fused.

\subsection{Custom Operation Kernel}
\label{subsec:custom_k}
Custom operation kernels are essential to optimize performance by eliminating computation overhead, e.g., unnecessary copying and intermediates. These kernel implementations must be mathematically equivalent to the reference code and can show significant performance gains under all or most circumstances, as discussed below.

\subsubsection{Theoretical deduction}
Instead of relying on the PyTorch implementation (Appendix ~\ref{subsect:ref}) of forward pass for focal loss and adopting the default generated backward pass, we implemented a custom focal loss kernel for both forward and backward pass (backward kernel implementation is optional, as implicit autograd can be generated). Focal loss can be represented as in Equation \ref{equ.focalloss} and we adopt $\gamma=2$ and $\alpha=0.25$. 

The forward pass can be simplified further by assuming x and y are real in Equation \ref{equ.myfocalloss}. Lastly, since y is a binary matrix, all the terms that contain \texttt{y(y-1)} equals to \texttt{0} and can be removed as shown in Equation \ref{equ.simpfocalloss}. The backward equation is shown in Appendix~\ref{subsec:focallossderiv}.

{\small
\begin{equation}
\label{equ.focalloss}
F L(p)=\left\{\begin{array}{cc}
-\alpha(1-p)^{\gamma} \log (p), & y=1 \\
-(1-\alpha) p^{\gamma} \log (1-p), & \text { otherwise }
\end{array}\right.
\end{equation}
}

{\small
\begin{equation}
\label{equ.myfocalloss}
F L=(a(2 y-1)-y+1)\left(\frac{-e^{x} y+e^{x}+y}{e^{x}+1}\right)^{\gamma}\left(\log \left(e^{x}+1\right)-x y\right)
\end{equation}
}

{\small
\begin{equation}
\label{equ.simpfocalloss}
F L_{sp}=\left(\frac{-e^{x} y+e^{x}+y}{e^{x}+1}\right)^{\gamma}\left((\alpha(2 y-1)-y+1) \log \left(e^{x}+1\right)-\alpha x y\right)
\end{equation}
}
\subsubsection{Implementation and Assessment}
The operators in ATEN of PyTorch can be roughly categorized into two types: in-place operation and standard operation, with the former suffixed by \_ (as in \textit{add\_}). Since in-place operation modifies the Tensor directly, the overhead of copying or creating new spaces in the cache is avoided. The implementation shown in Appendix~\ref{subsec:custom} heavily adopts in-place operation as much as possible, which enhances efficiency.

\begin{figure}[hpbt]
\centering
\includegraphics[width=0.75\linewidth]{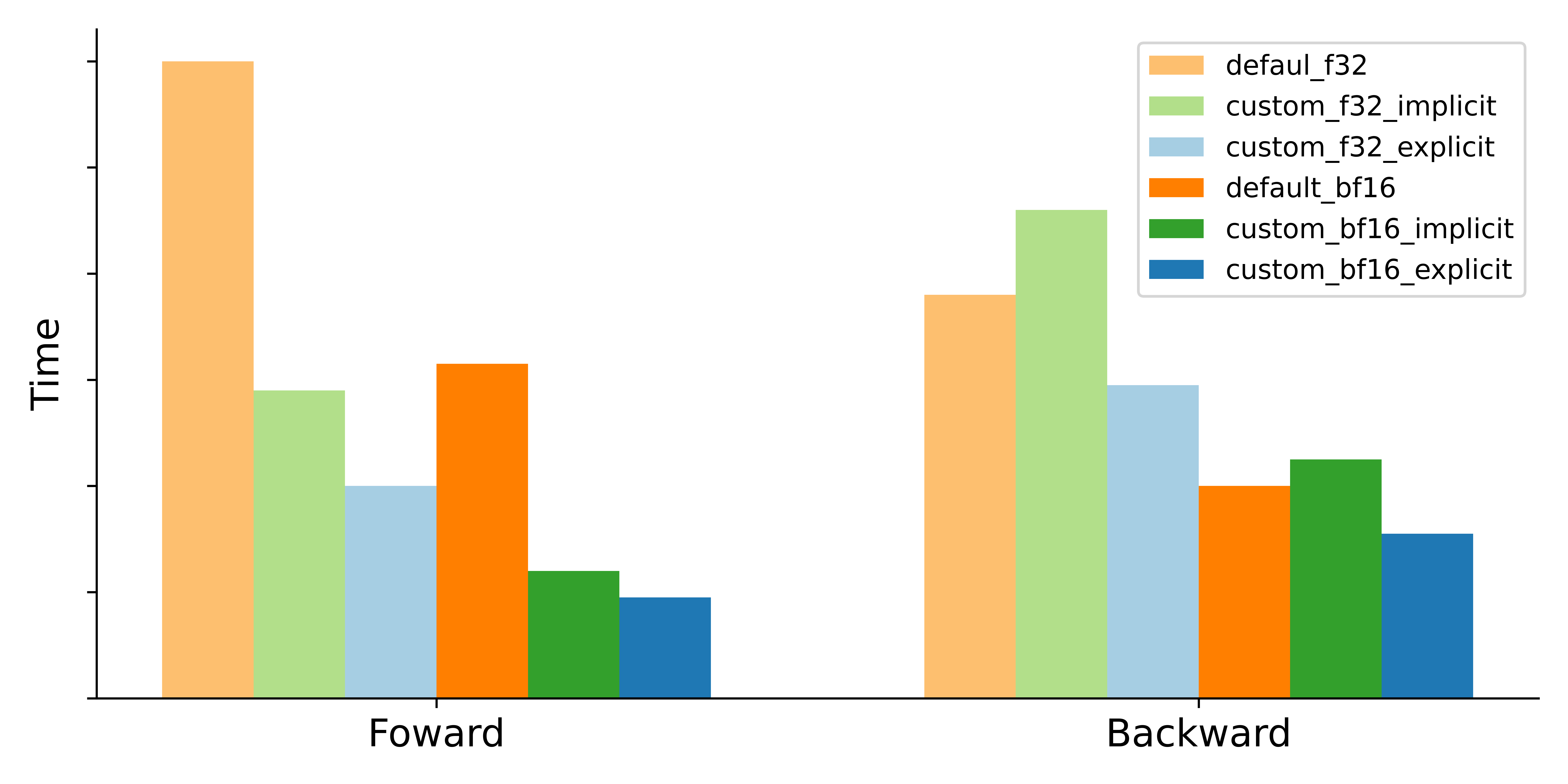}
\caption{Comparison of Custom Focal Loss Time vs Default}
\label{fig:kernel}
\end{figure}

After confirming that our kernel implementation is mathematical equivalent to the reference implementation, we tested our kernel against the reference code under both float32 and BF16 settings. As shown in Fig \ref{fig:kernel}, the custom forward kernel is 2.6 times faster than the default implementation under the BF16 setting. 

Although the PyTorch framework can generate implicit autograd for our custom kernel, its performance is not ideal. The custom backward kernel is 1.3 times faster than the reference implementation and 1.45 times faster than the generated implicit autograd kernel. We also discovered that the custom backward kernel can also boost forward kernel performance and we suspect that the explicit backward kernel can prevent the forward kernel from saving unnecessary intermediates. The combined improvement from the custom focal loss kernel is two times faster. Our code has been integrated into Intel\textsuperscript{\textregistered} Extension for PyTorch and will be available in that  library soon.

\section{Distributed Training}
\label{subsec:distributed}

Compared to inference, which can be scaled out amongst independent nodes, training DL models often require much greater computing power working synchronously. Meeting this need can be accomplished by scaling-up nodes with additional CPU resources or by scaling out amongst multiple nodes. When training a system at scale---whether multiple nodes, multiple sockets, or even a single socket---it is necessary to distribute the workload across multiple workers.

Coordination among distributed workers requires communication between them. Distributing workloads on CPUs can be performed via multiple protocols and middleware, such as MPI (Message Passing Interface)(\cite{gropp1999using}) and Gloo (\cite{gloo}). We use MPI terminology in subsequent sections.

\subsection{Distributed Training Performance}

To maximize training performance, a training workload should target one thread per CPU core of each system node. For example, an 8-socket system with 28 cores per socket should target 224 total threads. The total threads may be apportioned across several workers identified by their \textit{rank}, e.g., 8 ranks of 28 threads, 16 ranks of 14 threads, 32 ranks of 8 threads, etc. The selection of ranks and threads should not cause any rank to span multiple sockets. 

In practice, better performance may be achieved by utilizing more ranks with fewer threads each, rather than fewer ranks with more threads each at the same global batch size. Table \ref{table:scale} shows how the throughput goes up diagonally from bottom-left to top-right. 
\begin{table}[bhtp]
\centering
\begin{tabular}{l|lllll}
\hline 
 & \multicolumn{5}{c}{\textbf{Number of Workers}} \\
\hline 
\textbf{Threads/Worker} & \textbf{1} &  \textbf{2} & \textbf{4} & \textbf{8} & \textbf{16}
\\\hline 
\textbf{7} & 1.00 &  2.00 & 3.82 & 7.04 & 11.87
\\\hline 
\textbf{14} & 1.86 &  3.7 & 6.8 & 11.59 & -
\\\hline 
\textbf{28} & 3.27 &  6.51 & 11.20 & - & -
\\\hline 
\textbf{56} & 5.11 &  10.18 & - & - & -
\\\hline 
\end{tabular}
\caption{\label{table:scale} Scalability (Normalized Throughput) }
\end{table}
However, the number of available ranks is limited by the available system memory, model size, and batch size. The system memory is divided amongst the ranks, so each rank must have sufficient memory to support the model and host functions to avoid workload failure.

\subsection{Training Convergence}

As a training system is scaled-out to more nodes, sockets, or ranks, two factors are known to degrade the model’s convergence time: weak scaling efficiency and convergence point. Weak scaling efficiency is a ratio of \textit{the performance of a system to N systems doing N times as much work} and tends to lag behind the linear rate at which resources are added. This phenomenon and its causes are well-documented (\cite{sridharan2018scale}) across hardware types and are not explored further in this paper.

A model’s convergence point is the second factor that impacts  convergence time as a training system scales. In particular, the global batch size increases as a distributed system scales out, even though the local batch size per worker remains constant. For instance, if a 2-socket system launches a combined 8 ranks with a global batch size of 64 (BS=8 per rank), when scaled out to 8-sockets, the global batch size becomes 256 even though each rank has the same local batch size. 

\begin{figure}[hpbt]
\centering
\includegraphics[width=0.8\linewidth]{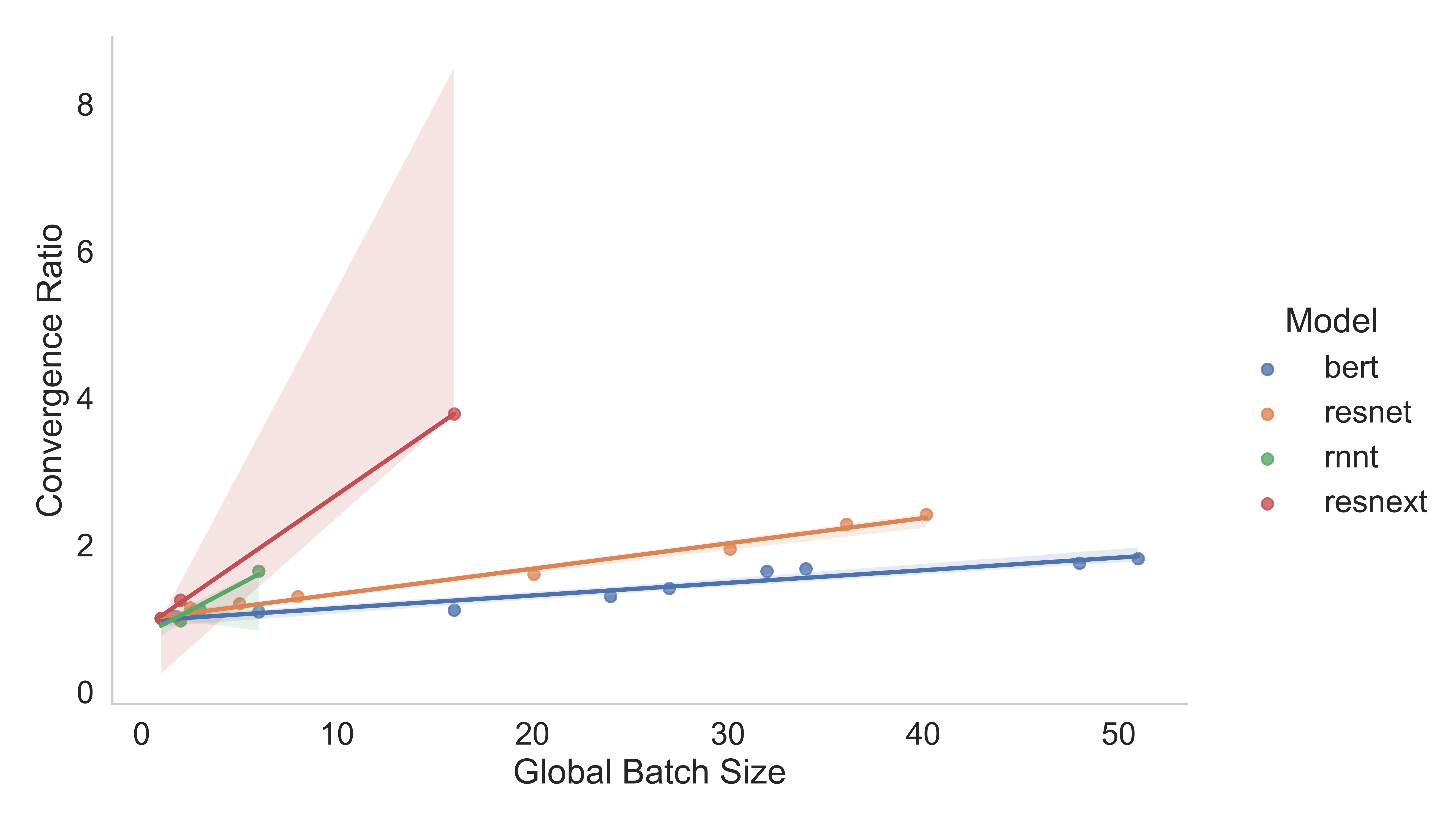}
\caption{Convergence Ratio vs Global Batch Size (Normalized)}
\label{fig:scale}
\end{figure}

As the number of epochs required to converge at a model’s target accuracy increases the global batch size of a training workload also increases, as shown in Fig \ref{fig:scale}. This increase in the epochs to reach a convergence point can detract substantially from the increased resources. When planning a system scale-out, it is therefore critical to account for the resulting convergence point and mitigates it by reducing the local batch size if possible (\cite{mlcommons}).

\section{Concluding Remarks}
\label{subsec:future}

This paper explores various aspects of optimization for training DL models on CPUs, in addition to a method guide. We present a DL profile/projection toolkit called ProfileDNN that helped us locate several issues for training RetinaNet-ResNext50, which when fixed, lead to a 2 times performance increase. We also created a custom Focal Loss kernel that is 1.5 times faster than the PyTorch reference implementation when running on CPUs. 

The following is a summary of the lessons learned from our
study of training deep learning models using CPUs:
\begin{itemize}
    \item Efficient DL frameworks that are optimized for CPUs like Intel\textsuperscript{\textregistered} extension for PyTorch can reduce training time dramatically with little cost.
    \item Model profiling should be done both on the reference code and custom implementations, especially when the data set is changed. Discrepancies between different implementations and corresponding low-level op distributions can help pinpoint the bottlenecks.
    \item Implementing both forward pass and backward pass explicitly for custom kernels leads to the best training performance.
    \item Local batch size is highly correlated with convergence point and should be reduced properly when planning a system scale-out.
\end{itemize}

Our future work will focus on testing our methodology and toolkit on other popular models (\cite{fu2022fastaudio}~\cite{fu2021transformer}) and conduct a more in-depth study on optimizing training DL models with distributed CPU clusters. As Large Language Models (LLMs) such as ChatGPT (\cite{white2023prompt}) gain widespread popularity, the significance of leveraging preexisting infrastructure grows more pronounced.

\acks{Thanks to Karen Perry for her feedback on this paper and her tireless support of the oneAPI middleware.}
\bibliography{acml22}

\appendix

\section{Appendix}

\subsection{Reference Focal Loss Code (\cite{marcel2010torchvision})}
\label{subsect:ref}
\begin{lstlisting}[language=Python]
import torch
import torch.nn.functional as F
import time
def sigmoid_focal_loss(
    inputs: torch.Tensor,
    targets: torch.Tensor,
    alpha: float = 0.25,
    gamma: float = 2,
    reduction: str = "none",
):
    inputs = inputs.to(dtype=torch.float32)
    targets = targets.to(dtype=torch.float32)
    p = torch.sigmoid(inputs)
    ce_loss = F.binary_cross_entropy_with_logits(
        inputs, targets, reduction="none"
    )
    p_t = p * targets + (1 - p) * (1 - targets)
    loss = ce_loss * ((1 - p_t) ** gamma)
    if alpha >= 0:
        alpha_t = alpha * targets + (1 - alpha) * (1 - targets)
        loss = alpha_t * loss
    if reduction == "mean":
        loss = loss.mean()
    elif reduction == "sum":
        loss = loss.sum()
    return loss
\end{lstlisting}

\subsection{Focal Loss Derivative}
\label{subsec:focallossderiv}
\begin{figure}[hpbt]
\centering
\includegraphics[width=0.6\linewidth]{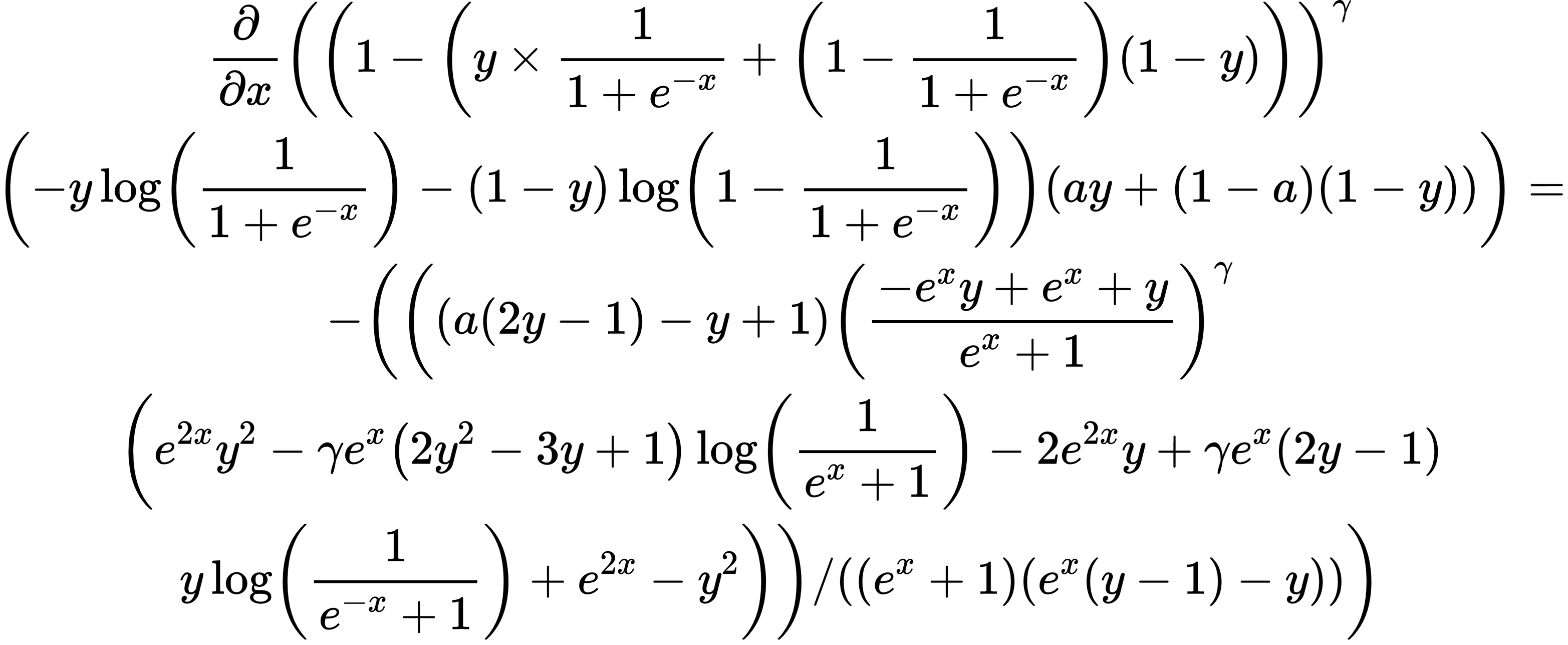}
\caption{Backward Kernel Equation}
\label{fig:focal_derive}
\end{figure}

\begin{figure}[hpbt]
\centering
\includegraphics[width=0.7\linewidth]{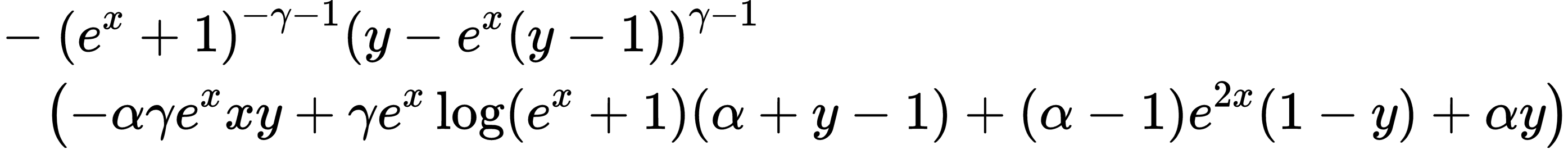}
\caption{Simplified Backward Kernel}
\label{fig:sip_back_kernel}
\end{figure}

\subsection{Custom Focal Loss Kernel Code}
\label{subsec:custom}
\begin{lstlisting}[language=C++]
at::Tensor _focal_loss_forward(const at::Tensor& input, const at::Tensor& target, const float alpha, const float gamma, const int64_t reduction) {
    at::Tensor loss;
    loss=(((alpha*(-input).mul_(target)).add_(((2*alpha-1)*target+(1-alpha)).mul_(((input.exp_() + 1).log_())))).mul_(((target -1).mul_(input).add_(-target)).pow_(gamma))).div_((input + 1).pow_(gamma));
    return apply_loss_reduction(loss, reduction);
}

at::Tensor _focal_loss_backward(const at::Tensor& grad, const at::Tensor& input, const at::Tensor& target, const float alpha, const float gamma, const int64_t reduction) {
    at::Tensor grad_input;
    grad_input=-((input.exp() + 1).pow(-gamma-1)).mul((target.add((1-target).mul(input.exp()))).pow(gamma - 1)).mul(((-alpha*gamma*input).mul(target).mul(input.exp())).add(gamma*(target+alpha-1).mul(input.exp()).mul(((input.exp()+1).log()))).add(alpha*target).add((alpha-1)*(1-target).mul((input.exp()).pow(2)))).mul(grad);
    if (reduction == at::Reduction::Mean) {
        return grad_input / input.numel();
    }
    return grad_input;
}
\end{lstlisting}

\end{document}